\documentclass[11pt]{article}
\usepackage{etex}
\usepackage[framed,numbered,autolinebreaks,useliterate]{mcode}
\usepackage{algorithm}
\usepackage{algpseudocode}

\usepackage[dvips]{hyperref}  
\hypersetup{
  colorlinks   = true, 
  urlcolor     = blue, 
  linkcolor    = blue, 
  citecolor   = red 
}
\usepackage[pageref]{backref}
\usepackage{url}
\usepackage{breakurl}
\usepackage{verbatim} 
\usepackage{hypernat}  
\usepackage{subeqnarray}
\usepackage{empheq} 
\usepackage{color} 
\usepackage{exscale}  
\usepackage{stmaryrd}  
\usepackage{epsfig}
\usepackage{subfigure}
\usepackage{cite}
\usepackage{calc}
\usepackage{times,amsfonts,amsmath,amssymb,amstext}
\usepackage{amsthm,amsbsy,bm}
\usepackage{mathrsfs}
\usepackage{dsfont} 
\usepackage{nicefrac}
\usepackage{CJK} %
\usepackage{multicol}
\usepackage{fancyhdr}
\usepackage[small]{caption}
\usepackage{empheq} 
\usepackage{color} 
\usepackage{picture,xcolor}
\usepackage{verbatim}
\usepackage{fancyvrb} 
\usepackage{enumerate}
\usepackage{cases}
\usepackage{varwidth}
\usepackage{exscale}  
\usepackage{datetime}
\usepackage{multirow,makecell,rotating}

\newfont{\boldit}{cmbxti10} 
\allowdisplaybreaks
\theoremstyle{definition}

\newtheorem{defi}{Definition}[section]

\theoremstyle{remark}

\usepackage[all]{xy}
\usepackage{xhfill}
\newcommand{\xfilll}[2][1ex]{%
  \dimen0=#2\advance\dimen0 by #1%
  \leaders\hrule height \dimen0 depth -#1\hfill%
}
\numberwithin{equation}{section}
\renewcommand*{\backref}[1]{} 
\renewcommand*{\backrefalt}[4]{%
    \ifcase #1 (Not cited.)%
    \or        (Cited on page~#2.)%
    \else      (Cited on pages~#2.)%
    \fi}

\usepackage{pgf,tikz}
\newcolumntype{L}{>{\varwidth[c]{\linewidth}}l<{\endvarwidth}}
\newcolumntype{M}{>{$}l<{$}}
\newcommand{\tc}[1]{\multicolumn{1}{c}{#1}} 
\newcommand{\tr}[1]{\multicolumn{1}{r}{#1}} 

\makeatletter
\def\hlinewd#1{%
\noalign{\ifnum0=`}\fi\hrule \@height #1 %
\futurelet\reserved@a\@xhline}

\makeatother

\makeatletter
\def\@cite#1#2{\textsuperscript{[{#1\if@tempswa , #2\fi}]}}
\makeatother

\makeatletter
  \def\my@tag@font{\normalsize}
  \def\maketag@@@#1{\hbox{\m@th\normalfont\my@tag@font#1}}
  \let\amsmath@eqref\eqref
  \renewcommand\eqref[1]{{\let\my@tag@font\relax\amsmath@eqref{#1}}}
\makeatother

\usepackage{siunitx}
\usepackage{pstricks-add}
\usepackage{siunitx}
\usepackage{pgfplots}
\usepackage{pgfplotstable}
\usepackage{filecontents}
\usepackage{graphicx}
\usepackage{latexsym}
\usepackage{keyval}
\usepackage{ifthen}
\usepackage{moreverb}
 \pgfplotsset{compat=newest}
 \usepackage{siunitx}

\usepackage{graphicx}
\usepackage{latexsym}
\usepackage{keyval}
\usepackage{ifthen}
\usepackage{moreverb}
\usepackage{gnuplottex}

\begin{document}

\title{\huge A Note on Newton-Like Iterative Solver for Multiple View L2 Triangulation}%
\author{{} {}\thanks{correspondence author}\\
}
\date{}
\maketitle

\begin{center}
{\bf Abstract}\\
\vspace{0.5cm}
\parbox{6.0in}
{\setlength
{\baselineskip}{0.5cm}

In this paper, we show that the L2 optimal solutions to most real multiple view L2 triangulation problems can be efficiently obtained by two-stage Newton-like iterative methods, while the difficulty of such problems mainly lies in how to verify the L2 optimality. Such a working two-stage bundle adjustment approach features the following three aspects: first, the algorithm is initialized by {\em symmedian point} triangulation, a multiple-view generalization of the mid-point method; second, a symbolic-numeric method is employed to compute derivatives accurately; third, globalizing strategy such as line search or trust region is smoothly applied to the underlying iteration which assures algorithm robustness in general cases.\\

Numerical comparison with {\em tfml} method shows that the {\em local} minimizers obtained by the two-stage iterative bundle adjustment approach proposed here are also the L2 optimal solutions to all the calibrated data sets available online by the Oxford visual geometry group. Extensive numerical experiments indicate the bundle adjustment approach solves more than 99\% the real triangulation problems optimally. An IEEE 754 double precision C++ implementation shows that it takes only about 0.205 second to compute all the 4983 points in the Oxford {\em dinosaur} data set via Gauss-Newton iteration hybrid with a line search strategy on a computer with a 3.4GHz Intel\textsuperscript{\textregistered} i7 CPU.
} \end{center}
\vspace{0.5 cm}
\section{Introduction}

Triangulation is a critical topic in computer vision with applications in 3D object reconstruction, map estimation, robotic path-planning, surveillance and virtual reality~\cite{Hartley97d, Hartley2013IJCV, Chesi2011tfml, Recker2013WACV}. Efficient two-view triangulation methods~\cite{Lindstrom2010, Wu2011a} and especially multiple-view L2 optimal ones~\cite{HowHard3view2005,Lu2007,Byroed07,Chesi2011tfml} have drawn intensive research interests; the latter give rise to favorable maximum likelihood estimates under the assumption of independent gaussian noises~\cite{Hartley97d} but still remain not well-resolved.%

Triangulation algorithms which guarantee L2 optimality for up to three-view cases are mainly based on polynomial solving, symbolic-numeric Gr\"{o}bner basis methods in solving polynomial systems, and branch-and-bounds optimization techniques~\cite{Hartley97d, HowHard3view2005, Lu2007, Byroed07}. Recent research indicates that such an algorithm as can find a closed-form $n$-view L2 optimal solution does not exist~\cite{Hartley2013IJCV}. %

A novel non-iterative method based on fundamental matrix and linear matrix inequalities, {\em tfml}, by Chesi et al~\cite{Chesi2011tfml}, is efficient and able to handle more than three-view L2 triangulation. The major limitations of {\em tfml} might be the low solution accuracy in the {\em conservative cases}~\cite{Chesi2011tfml} and the fast efficiency decline due to scale increasing of the converted eigen value problem(EVP) when the number of cameras increases. Despite of these, {\em tfml} is probably by far the most successful $n$-view L2 triangulation method created naturally with a {\em necessary and sufficient} cirtierion for L2 optimality verification~\cite{Chesi2011tfml,Hartley2013IJCV} and will be used as benchmark here.

Traditional iterative methods such as the bundle adjustment optimization via Levenberg-Marquardt are mainly criticized for their no ideal initialization and the possible local convergence issue~\cite{Triggs:1999iccv, ZissermanHartley-5, HowHard3view2005, Lu2007, Byroed07, Hartley2013IJCV}. As far as we know, none of the state-of-the-art triangulation approaches which asserts L2 optimality for multiple view triangulation are iterative methods. Recent publications indicate that {\em bundle adjustment} optimization performs poorer than even some of the suboptimal methods~\cite{Recker2013WACV, Recker2013swarm}.

We find most of the real $n$-view L2 triangulation problems don't have the difficulty of multiple local minima, i.e., in most cases the global L2 optimal solutions can be approached by solving only a convex problem via simple Newton-like methods~\cite{Hartley2013IJCV}. As a matter of fact, a lot of the most cited real data sets can be globally solved by iterative methods with excellent accuracy and high efficiency. These data sets include but are not limited to {\em dinosaur, model house, corridor, Merton colleges I, II} and {\em III, University library} and {\em Wadham College}, which are made available online by the visual geometry group of Oxford university(VGG, \url{http://www.robots.ox.ac.uk/~vgg/data/data-mview.html}) and are widely used to evaluate new triangulation algorithms.

In our numerical experiments, Newton-Raphson, Gauss-Newton and Levenberg-Marquardt methods all work successfully on Oxford VGG data when being implemented by:
\vspace{-0.1 cm}
\begin{enumerate}[(1)]
\item initializing via {\em symmedian-point} triangulation to obtain a good start point;
\vspace{-0.3 cm}
\item computing all derivatives, gradients and Hessians of the cost function included, via a symbolic-numeric approach (or multiple precision computation) to assure high accuracy;
\vspace{-0.3 cm}
\item using Newton-like underlying iterative methods such as Newton-Raphson, Gauss-Newton and other variants, smoothly hybrid with globalizing strategies in order to handle hard cases when symmedian point is not a good start point.
\end{enumerate}
\vspace{-0.1 cm}

In this work we will show these implementation details and briefly introduce some criteria useful in verifying the L2 optimality~\cite{wedin:1977,Hartley2013IJCV,Chesi2011tfml}. We intend to present that bundle adjustment optimization with appropriate implementation details is a practically well-performed approach in solving the multiple-view L2 triangulation problems.

\section{Implementation details of the iterative solver}\label{cost function equivalent section}

The cost function of a typical unconstrained least square problem has the following form~\cite{Nocedal2006}:
\begin{equation}\label{least square cost function}
\dfrac{1}{2} f\left(X\right) = \dfrac{1}{2}  r\left(X\right)^T\; r\left(X\right) = \dfrac{1}{2} \sum\limits_{i = 1}^{m} {\phi _i^2\left( {X} \right)}
\end{equation}

The $n$-view L2 triangulation is the least square problem as in~\eqref{reprojection error cost function}: given $n$ pinhole cameras ${\boldsymbol P}_i$ in 3$\times$4 and $n$ 2D image homogenous coordinates ${\bf x}_i=\left( u_i,v_i,1 \right)^T$, find the global least square minimizer $X^*$:
\begin{equation}\label{reprojection error cost function}
X^* =  \mathop{\text{arg}}\mathop{\min}\limits_{X\in{\mathbb{R}^3}} f\left( X \right) = \mathop{\arg}\mathop{\min}\limits_{X\in{\mathbb{R}^3}}\sum\limits_{i = 1}^n {\left\| {\bf x}_i - \hat{\bf x}_i \right\|_2^2}, \quad \text{where}:
\hat{\bf x}_i = {\left( \hat{u}_i,\hat{v}_i,1 \right)^T} = \dfrac{{\boldsymbol P}_i}{\lambda _i}X^\minuso, i=1\cdots n.
 \end{equation}
\noindent $X=(x,y,z)^T$ represents a 3D scene point, $X^\minuso = \left(x,y,z,1 \right)^T$ is $X$ in homogeneous coordinates and $\lambda_i$ is the projective depth corresponding to ${\boldsymbol P}_i$. 

The projection of $n$-camera cases can be represented as in equation~\eqref{nViewProjection}:

\begin{equation}\label{nViewProjection}\left[ {\begin{array}{*{20}{c}}
{\begin{array}{*{20}{c}}
{{\lambda _1} \; {u_1}}\\
{{\lambda _1} \; {v_1}}\\
{{\lambda _1}}
\end{array}}\\[20pt]
{\begin{array}{*{20}{c}}
{{\lambda _2} \; {u_2}}\\
{{\lambda _2} \; {v_2}}\\
{{\lambda _2}}
\end{array}}\\
 \vdots \\
{\begin{array}{*{20}{c}}
{{\lambda _n} \; {u_n}}\\
{{\lambda _n} \; {v_n}}\\
{{\lambda _n}}
\end{array}}
\end{array}} \right] = \left[ {\begin{array}{*{20}{c}}
{\begin{array}{*{20}{c}}
{p_{11}^1}&{p_{12}^1}&{p_{13}^1}&{p_{14}^1}\\
{p_{21}^1}&{p_{22}^1}&{p_{23}^1}&{p_{24}^1}\\
{p_{31}^1}&{p_{32}^1}&{p_{33}^1}&{p_{34}^1}
\end{array}}\\[20pt]
{\begin{array}{*{20}{c}}
{p_{11}^2}&{p_{12}^2}&{p_{13}^2}&{p_{14}^2}\\
{p_{21}^2}&{p_{22}^2}&{p_{23}^2}&{p_{24}^2}\\
{p_{31}^2}&{p_{32}^2}&{p_{33}^2}&{p_{34}^2}
\end{array}}\\
 \vdots \\
{\begin{array}{*{20}{c}}
{p_{11}^n}&{p_{12}^n}&{p_{13}^n}&{p_{14}^n}\\
{p_{21}^n}&{p_{22}^n}&{p_{23}^n}&{p_{24}^n}\\
{p_{31}^n}&{p_{32}^n}&{p_{33}^n}&{p_{34}^n}
\end{array}}
\end{array}} \right]\left[ {\begin{array}{*{20}{c}}
x\\
y\\
z\\
1
\end{array}} \right]\end{equation}
\noindent and the $2n\times 1$ residue vector $r(X)$ of~\eqref{reprojection error cost function} can be written as in~\eqref{residue vector function}:
\begin{equation}\label{residue vector function}
r\left(X\right) = \left[ {\begin{array}{*{20}{c}}
{\begin{array}{*{20}{c}}
{p_{11}^{*1}}&{p_{12}^{*1}}&{p_{13}^{*1}}\\
{p_{21}^{*1}}&{p_{22}^{*1}}&{p_{23}^{*1}}
\end{array}}\\[15pt]
{\begin{array}{*{20}{c}}
{p_{11}^{*2}}&{p_{12}^{*2}}&{p_{13}^{*2}}\\
{p_{21}^{*2}}&{p_{22}^{*2}}&{p_{23}^{*2}}
\end{array}}\\
 \vdots \\
{\begin{array}{*{20}{c}}
{p_{11}^{*n}}&{p_{12}^{*n}}&{p_{13}^{*n}}\\
{p_{21}^{*n}}&{p_{22}^{*n}}&{p_{23}^{*n}}
\end{array}}
\end{array}} \right]\left[ {\begin{array}{*{20}{c}}
x\\
y\\
z
\end{array}} \right] - \left[ {\begin{array}{*{20}{c}}
{\begin{array}{*{20}{c}}
{{u_1} - p_{14}^{*1}}\\
{{v_1} - p_{24}^{*1}}
\end{array}}\\[15pt]
{\begin{array}{*{20}{c}}
{{u_2} - p_{14}^{*2}}\\
{{v_2} - p_{24}^{*2}}
\end{array}}\\
 \vdots \\
{\begin{array}{*{20}{c}}
{{u_n} - p_{14}^{*n}}\\
{{v_n} - p_{24}^{*n}}
\end{array}}
\end{array}} \right] = A X - B = \left[ {\begin{array}{*{20}{c}}
\begin{array}{*{20}{c}}
\phi _1 \left( X \right)\\
\phi _2 \left( X \right)
\end{array}\\[15pt]
\begin{array}{*{20}{c}}
\phi _3 \left( X \right)\\
\phi _4 \left( X \right)
\end{array}\\
 \vdots \\
{\begin{array}{*{20}{c}}
\phi _{2n-1}\left( X \right)\\
\phi _{2n}\left( X \right)
\end{array}}
\end{array}} \right] \end{equation}
\noindent where $p_{l,m}^{*i}= \dfrac{p_{l,m}^{i}}{\lambda_i} \left(\lambda_i = p_{3,1}^i x+p_{3,2}^i y + p_{3,3}^i z + p_{3,4}^i, i= 1, 2, \cdots, n, l=1,\cdots,3, m=1,\cdots,4 \right)$.

Then cost function $f(X)$ in~\eqref{reprojection error cost function} has the equivalent least square problem form as in equation~\eqref{least square cost function}~\cite{Nocedal2006} with $m=2n$. 

\subsection{Symmedian point method for initialization}\label{suboptimal methods section}

\noindent The fast two-view mid-point triangulation method~\cite{Hartley97d, ZissermanHartley-5} has been extended to $n$-view cases by generalizing the concept {\em mid-point} into {\em symmedian point} which has the least sum of squared distances to all the projection rays. This idea was initially proposed by Sturm et al in 2006~\cite{sturm:midpoint}, a simple and detailed implementation of which can also be found in~\cite[pp.305~\texttildelow~307] {Szeliski:ComputerVision2011}. However, it seems the advantage of such method in initializing an L2 triangulation has not yet been sufficiently realized.

In 3D Euclidean space, a line $l_i$ can be defined by a fixed point $S_i$ and a direction $W_i$ as:
\begin{equation}
{l_i} \triangleq \left\langle {{S_i},{W_i}} \right\rangle
\end{equation}
where $W_i$ is a 3$\times$1 unit direction vector with 2-norm equal to 1.  Define the 3$\times$3 projection $P_i$ as:
\begin{equation}
{P_i} \triangleq I_{3\times 3} - \frac{{{W_i} \; W_i^T}}{{W_i^T \; {W_i}}} = I_{3\times 3} - {W_i} \; W_i^T
\end{equation}
because the distance $d_i$ between $X$ and line $l_i$ $= \left\langle {{S_i},{W_i}} \right\rangle $ satisfies the quadratic form:
\begin{equation}\label{symmedian quadratic form}
{d_i^2} = \left\| P_i\left( X - S_i \right) \right\|_2^2 =\left( X - S_i \right)^T \left( P_i^T P_i\right) \left( X - S_i \right)
\end{equation}
then the {\em symmedian point} $\hat X$ which minimizes the sum of the $n$ quadratic forms by~\eqref{symmedian quadratic form} can be obtained by solving the $3\times 3$ linear system of equations~\eqref{symmedian point method linear equations}:
\begin{equation}\label{symmedian point method linear equations}
\left( {\sum\limits_{i = 1}^n {{P_i}} } \right) \; \hat{X} = \sum\limits_{i = 1}^n {\left( {{P_i} \; {S_i}} \right)} \end{equation}

This triangulation method requires pinhole camera factorization such that all projection rays can be represented into a fixed 3D point $S_i$  and its direction $W_i$ both in their Euclidean coordinates~\cite{Szeliski:ComputerVision2011}. Such a multiple-view triangulation approach via {\em symmedian point} is linear, suboptimal and efficient. The {\em symmedian points} thus obtained in closed form usually are excellent initial values for further improvement in those two-stage triangulation methods~\cite{Byroed07, Recker2013WACV}.

 Note that the two-stage iterative methods find the L2 optimal solutions only when the initial triangulation locates the global-L2-optimal attaction basin of the problems correctly. However, it is difficult to clarify which initialization algorithm is in general better than others. It is in our extensive numerical experiments that we find {\em symmedian point} triangulation outperforms other linear triangulation methods and comparison details are omitted here. The iterative methods discussed in this work are all initialized by {\em symmedian points}, while the {\em tfml} method by Chesi et al~\cite{Chesi2011tfml} also works but is too much expensive.

\subsection{Symbolic-numeric computation of derivatives}\label{section:symbolic-numeric derivatives}

 A symbolic-numeric approach is employed to compute accurate derivatives of~\eqref{reprojection error cost function}. It because the subtle changes in the implementation of Newton-like iterative methods may causes significant difference in the numerical solutions to the general multiple-view triangulation why we present these implementation details in a separate subsection. In fact, this is probably one of the reasons why bundle adjustment optimization has long been considered as at most suboptimal even for Oxford VGG data sets besides no good initialization and the absence of optimality verification criterion~\cite{Hartley2013IJCV}.

Denote the $2n\times 3$ dimensional Jacobian matrix of $r\left(X\right)$~\eqref{residue vector function} as:
\begin{equation}\label{Jacobian of residue function}
J\left( X \right) = \left[ {\begin{array}{*{20}{c}}
{\dfrac{{\text{\scalebox{1}[0.85]{\rotatebox{10}{$\partial$}}} {\phi _1}}}{{\partial x}}}&{\dfrac{{\partial {\phi _1}}}{{\partial y}}}&{\dfrac{{\partial {\phi _1}}}{{\partial z}}}\\[12pt]
{\dfrac{{\partial {\phi _2}}}{{\partial x}}}&{\dfrac{{\partial {\phi _2}}}{{\partial y}}}&{\dfrac{{\partial {\phi _2}}}{{\partial z}}}\\[12pt]
{}& \vdots &{}\\[12pt]
{\dfrac{{\partial {\phi _{2n}}}}{{\partial x}}}&{\dfrac{{\partial {\phi _{2n}}}}{{\partial y}}}&{\dfrac{{\partial {\phi _{2n}}}}{{\partial z}}}%
\end{array}} \right]_{2n\times 3}\end{equation}

Then the gradient $g$ and Hessian $H$ of the cost function $f(X)$ in~\eqref{reprojection error cost function} can be represented as~\cite{Nocedal2006}:
\begin{equation}\label{gradient of fX}
g(X)\;{\buildrel \Delta \over =}\; \nabla f(X) = J(X)^T r(X) = J(X)^T \left(AX-B\right)
\end{equation}
\begin{equation}\label{Hessian of fX}
H(X) \;{\buildrel \Delta \over =}\; \nabla^2 f(X) = J(X)^T J(X) + \sum\limits_{i=1}^{2n}{\phi_i \nabla^2\phi_i}
\end{equation}

It is critical to assure appropriately high accuracy of $g(X)$~\eqref{gradient of fX} since the least square problem is converted into solving a nonlinear system~\eqref{nonlinear normal equation} using iterative methods. Any perturbation on $g(X)$, numerical round-off errors included, means using solutions of a perturbed system $\hat g(X)=0$ to approximate that of~\eqref{nonlinear normal equation}. Inappropriate approximation to derivatives may be one of the reasons why conventional implementation of iterative methods work unsatisfactory even for L2 triangulation problems close to the noise-free trivial cases.

\begin{equation}\label{nonlinear normal equation}
g \left(X\right)= J\left( X \right)^T\; r\left( X\right) = J\left( X \right)^T\; \left(A\; X -B \right) = \bf{0}
\end{equation}

The gradient~\eqref{gradient of fX} and Hessian~\eqref{Hessian of fX} of reprojection error cost function~\eqref{reprojection error cost function} can be accurately estimated via a symbolic-numeric approach.

Considering the $i$-th partition of $r(X)$ as in~\eqref{residue vector function}, which consists of $\phi_{2i-1}$ and $\phi_{2i}$ corresponding to the $i$-th camera:
\begin{equation}\label{residue vector function ith partition}
r_i\left(X\right)=\left( {\begin{array}{*{20}{c}}
{{\phi_{2i - 1}}}\\[12pt]
{{\phi_{2i}}}
\end{array}} \right) = \left( {\begin{array}{*{20}{c}}
{\dfrac{{{{p}}_{11}^ix + {{p}}_{12}^iy + {{p}}_{13}^iz + {{p}}_{14}^i}}{{{{p}}_{31}^ix + {{p}}_{32}^iy + {{p}}_{33}^iz + {{p}}_{34}^i}} - {u_i}}\\[12pt]
{\dfrac{{{{p}}_{21}^ix + {{p}}_{22}^iy + {{p}}_{23}^iz + {{p}}_{24}^i}}{{{{p}}_{31}^ix + {{p}}_{32}^iy + {{p}}_{33}^iz + {{p}}_{34}^i}} - {v_i}}
\end{array}} \right)\end{equation}
\noindent Since both $\phi_{2i-1}$ and $\phi_{2i}$ are in rational forms and all such partitions are independent from each other, the first and second order partial derivatives of them and therefore the $J$~\eqref{Jacobian of residue function}, $g$~\eqref{gradient of fX} and $H$~\eqref{Hessian of fX} can all be computed in a symbolic-numeric manner accurately.

For example, the $i$-th partition of $J(X)$ is:
\begin{equation}\label{Jacobian ith partition}
J_i\left(X\right)=\left( {\begin{array}{*{20}{c}}
{{J_{2i - 1}}}\\[12pt]
{{J_{2i}}}
\end{array}} \right) = \left[ {\begin{array}{*{20}{c}}
{\dfrac{{\partial {\phi_{2i - 1}}}}{{\partial x}}}&{\dfrac{{\partial {\phi_{2i - 1}}}}{{\partial y}}}&{\dfrac{{\partial {\phi_{2i - 1}}}}{{\partial z}}}\\[12pt]
{\dfrac{{\partial {\phi_{2i}}}}{{\partial x}}}&{\dfrac{{\partial {\phi_{2i}}}}{{\partial y}}}&{\dfrac{{\partial {\phi_{2i}}}}{{\partial z}}}
\end{array}} \right]\end{equation}

Denote $\lambda_i = p_{3,1}^i x+p_{3,2}^i y + p_{3,3}^i z + p_{3,4}^i$ and the following 18 determinants as:

\begin{eqnarray}\label{18 determinants}
\hphantom{a}1) \Delta _{12}^{i,1} = p_{11}^ip_{32}^i - p_{12}^ip_{31}^i,\hphantom{aa}2)\Delta _{13}^{i,1} = p_{11}^ip_{33}^i - p_{13}^ip_{31}^i,\hphantom{aa}3)\Delta _{14}^{i,1} = p_{11}^ip_{34}^i - p_{14}^ip_{31}^i,\nonumber\\[8pt]
\hphantom{a}4)\Delta _{21}^{i,1} = p_{12}^ip_{31}^i - p_{11}^ip_{32}^i,\hphantom{aa}5)\Delta _{23}^{i,1} = p_{12}^ip_{33}^i - p_{13}^ip_{32}^i,\hphantom{aa}6)\Delta _{24}^{i,1} = p_{12}^ip_{34}^i - p_{14}^ip_{32}^i,\nonumber\\[8pt]
\hphantom{a}7)\Delta _{31}^{i,1} = p_{13}^ip_{31}^i - p_{11}^ip_{33}^i,\hphantom{aa}8)\Delta _{32}^{i,1} = p_{13}^ip_{32}^i - p_{12}^ip_{33}^i,\hphantom{aa}9)\Delta _{34}^{i,1} = p_{13}^ip_{34}^i - p_{14}^ip_{33}^i,\nonumber\\[8pt]
10)\Delta _{12}^{i,2} = p_{21}^ip_{32}^i - p_{22}^ip_{31}^i,\hphantom{a}11)\Delta _{13}^{i,2} = p_{21}^ip_{33}^i - p_{23}^ip_{31}^i,\hphantom{a}12)\Delta _{14}^{i,2} = p_{21}^ip_{34}^i - p_{24}^ip_{31}^i,\nonumber\\[8pt]
13)\Delta _{21}^{i,2} = p_{22}^ip_{31}^i - p_{21}^ip_{32}^i,\hphantom{a}14)\Delta _{23}^{i,2} = p_{22}^ip_{33}^i - p_{23}^ip_{32}^i,\hphantom{a}15)\Delta _{24}^{i,2} = p_{22}^ip_{34}^i - p_{24}^ip_{32}^i,\nonumber\\[8pt]
16)\Delta _{31}^{i,2} = p_{23}^ip_{31}^i - p_{21}^ip_{33}^i,\hphantom{a}17)\Delta _{32}^{i,2} = p_{23}^ip_{32}^i - p_{22}^ip_{33}^i,\hphantom{a}18)\Delta _{34}^{i,2} = p_{23}^ip_{34}^i - p_{24}^ip_{33}^i \hphantom{a}
\end{eqnarray}

\noindent let the Kronecker product of $3\times 3$ identity matrix and $X^{\minuso}=\left(x,y,z,1\right)^T$ be:
\begin{equation}\label{kronecker product of X}
\text{Kron}\left( X^{\minuso} \right)
 = \left( {\begin{array}{*{20}{c}}
1&0&0\\
0&1&0\\
0&0&1
\end{array}} \right) \otimes \left[ {\begin{array}{*{20}{c}}
x\\
y\\
z\\
1
\end{array}} \right] = {\left( {\begin{array}{*{20}{c}}
X^{\minuso}&\bf 0&\bf 0\\
\bf 0&X^{\minuso}&\bf 0\\
\bf 0&\bf 0&X^{\minuso}
\end{array}} \right)_{12 \times 3}}
\end{equation}
\noindent and the numerical part $J_{\text{num}}^i$ of Jacobian's $i$-th partition independent of any variable $x$, $y$ or $z$ be:
\begin{equation}\label{numerical Jacobian ith partition}
J_{\text{num}}^i=\left( {\begin{array}{*{20}{c}}
0&{\Delta _{12}^{i,1}}&{\Delta _{13}^{i,1}}&{\Delta _{14}^{i,1}}&{\Delta _{21}^{i,1}}&0&{\Delta _{23}^{i,1}}&{\Delta _{24}^{i,1}}&{\Delta _{31}^{i,1}}&{\Delta _{32}^{i,1}}&0&{\Delta _{34}^{i,1}}\\[12pt]
0&{\Delta _{12}^{i,2}}&{\Delta _{13}^{i,2}}&{\Delta _{14}^{i,2}}&{\Delta _{21}^{i,2}}&0&{\Delta _{23}^{i,2}}&{\Delta _{24}^{i,2}}&{\Delta _{31}^{i,2}}&{\Delta _{32}^{i,2}}&0&{\Delta _{34}^{i,2}}
\end{array}} \right)\end{equation}
\noindent then $J(X)$'s $i$-th partition $J_i(X)$ as in~\eqref{Jacobian ith partition} can be represented by:
\begin{equation}\label{Jacobian ith partition new}
J_i\left(X\right) =\left( {\begin{array}{*{20}{c}}
{{J_{2i - 1}}}\\[12pt]
{{J_{2i}}}
\end{array}} \right) = J_{\text{num}}^i\; \text{Kron}\left(X^{\minuso}\right)*\lambda^{-2}_i
\end{equation}

Note that no such numerical approximation as finite difference is needed when calculating the $J$~\eqref{Jacobian of residue function} and then $g$~\eqref{gradient of fX} this way. The numerical parts $J_{\text{num}}^i$'s~\eqref{numerical Jacobian ith partition} of $J_i(X)$'s~\eqref{Jacobian ith partition new} can be pre-calculated since they only depend on the cameras. It is already enough with only the accurate $J$~\eqref{Jacobian of residue function} and $g$~\eqref{gradient of fX} in the Gauss-Newton and Levenberg-Marquardt methods~\cite{Nocedal2006} where second order derivatives are unnecessary.

Accurate Hessian $H$~\eqref{Hessian of fX} can also be obtained based on the accurate $J$ per~\eqref{Jacobian ith partition new} and the analytical second order derivatives of~\eqref{residue vector function ith partition} in similar way; and the first order finite-difference approximation to $H$~\eqref{Hessian of fX}  based on the accurate $J$~\eqref{Jacobian of residue function} and $g$~\eqref{gradient of fX} also works well. 

Per~\eqref{Hessian of fX}, we only need to further compute $\nabla^2\phi_{2i-1}$ and $\nabla^2\phi_{2i}$. Since both $\nabla^2\phi_{2i-1}$ and $\nabla^2\phi_{2i}$ are $3\times 3$ symmetric matrices, each of them has only 6 independent entries. Number the 12 entries in the sequence as defined in~\eqref{eqn:indices of delta2phi_i}, then every 6 of them can be rewritten into a $6\times 1$ vector:
\begin{equation}\label{eqn:indices of delta2phi_i}
\text{indices of }\nabla^2\phi_{2i-1}%
: \left[
\begin{array}{ccc}
1 & 2 & 3\\
2 & 4 & 5\\
3 & 5 & 6\\
\end{array}\right] \mapsto
\left[
\begin{array}{c}
1\\
2\\
3\\
4\\
5\\
6\\
\end{array}\right];
\quad\text{indices of }\nabla^2\phi_{2i}
:\left[
\begin{array}{ccc}
7 & 8 & 9\\
8 & 10&11\\
9 & 11&12\\
\end{array}\right] \mapsto
\left[
\begin{array}{c}
7\\
8\\
9\\
10\\
11\\
12\\
\end{array}\right]
\end{equation}

All the 12 independent entries of $\nabla^2\phi_{2i-1}$ and $\nabla^2\phi_{2i}$ for the $i$-th camera can be concatenated together as one $12\times 1$ dimensional column vector $h_{12}^i$, then be represented by the product of the following $12\times 4$ matrix $H_{\text{num}}^i$ and the homogeneous vector $(x,y,z,1)^T*\lambda_i^{-3}$ $\left(\text{where: }\lambda_i = p^i_{31}x+p^i_{32}y+p^i_{33}z+p^i_{34}\right)$ :
\begin{equation}\label{numerical Hessian ith partition}
H_{\text{num}}^i =
\left(
\begin{array}{cccc}
0&2{p^i_{31}}\Delta^{i,1}_{21}&2{p^i_{31}}\Delta^{i,1}_{31}&2{p^i_{31}}\Delta^{i,1}_{41}\\[8pt]
{p^i_{31}}\Delta^{i,1}_{12}&{p^i_{32}}\Delta^{i,1}_{21}&p^i_{21}\Delta^{i,1}_{32}+p^i_{32}\Delta^{i,1}_{31}&p^i_{31}\Delta^{i,1}_{42}+p^i_{32}\Delta^{i,1}_{41}\\[8pt]
{p^i_{31}}\Delta^{i,1}_{13}&p^i_{21}\Delta^{i,1}_{23}+p^i_{33}\Delta^{i,1}_{21}&{p^i_{33}}\Delta^{i,1}_{31}&p^i_{31}\Delta^{i,1}_{43}+p^i_{33}\Delta^{i,1}_{14}\\[8pt]
2{p^i_{32}}\Delta^{i,1}_{12}&0&2{p^i_{32}}\Delta^{i,1}_{32}&2{p^i_{32}}\Delta^{i,1}_{42}\\[8pt]
p^i_{33}\Delta^{i,1}_{12}+p^i_{32}\Delta^{i,1}_{13}&{p^i_{32}}\Delta^{i,1}_{23}&{p^i_{33}}\Delta^{i,1}_{32}&p^i_{32}\Delta^{i,1}_{43}+p^i_{33}\Delta^{i,1}_{42}\\[8pt]
2{p^i_{33}}\Delta^{i,1}_{13}&2{p^i_{33}}\Delta^{i,1}_{23}&0&2{p^i_{33}}\Delta^{i,1}_{43}\\[8pt]
0&2{p^i_{31}}\Delta^{i,2}_{21}&2{p^i_{31}}\Delta^{i,2}_{31}&2{p^i_{31}}\Delta^{i,2}_{41}\\[8pt]
{p^i_{31}}\Delta^{i,2}_{12}&{p^i_{32}}\Delta^{i,2}_{21}&p^i_{21}\Delta^{i,2}_{32}+p^i_{32}\Delta^{i,2}_{31}&p^i_{31}\Delta^{i,2}_{42}+p^i_{32}\Delta^{i,2}_{41}\\[8pt]
{p^i_{31}}\Delta^{i,2}_{13}&p^i_{21}\Delta^{i,2}_{23}+p^i_{33}\Delta^{i,2}_{21}&{p^i_{33}}\Delta^{i,2}_{31}&p^i_{31}\Delta^{i,2}_{43}+p^i_{33}\Delta^{i,2}_{14}\\[8pt]
2{p^i_{32}}\Delta^{i,2}_{12}&0&2{p^i_{32}}\Delta^{i,2}_{32}&2{p^i_{32}}\Delta^{i,2}_{42}\\[8pt]
p^i_{33}\Delta^{i,2}_{12}+p^i_{32}\Delta^{i,2}_{13}&{p^i_{32}}\Delta^{i,2}_{23}&{p^i_{33}}\Delta^{i,2}_{32}&p^i_{32}\Delta^{i,2}_{43}+p^i_{33}\Delta^{i,2}_{42}\\[8pt]
2{p^i_{33}}\Delta^{i,2}_{13}&2{p^i_{33}}\Delta^{i,2}_{23}&0&2{p^i_{33}}\Delta^{i,2}_{43}\\[8pt]
\end{array}
\right)
\end{equation}
All the $\Delta^{i,l}_{mn}$'s are as those defined in equation~\eqref{18 determinants}. Then we can obtain accurate Hessian of $f(X)$~\eqref{reprojection error cost function} per~\eqref{Hessian of fX}. Note that $\Delta_{12}^{i,l}=-\Delta_{21}^{i,l}$, $\Delta_{13}^{i,l}=-\Delta_{31}^{i,l}$, $\Delta_{23}^{i,l}=-\Delta_{32}^{i,l}$, $\forall l=1,2$.

\subsection{Newton-like iterative methods and the globalizing strategies}\label{section:iterative methods}
Many state-of-the-art nonlinear optimizers can be used to minimize the unconstrained $f(X)$~\eqref{reprojection error cost function}. The classical Newton-like iterative methods, Newton-Raphson, Gauss-Newton and Levenberg-Marquardt methods, have locally superlinear and quadratic convergence rate~\cite{iterative1970, Dennis:1996, Nocedal2006, NumericalRecipes2007a, NewtonIteration2008, NewtonAffineInvariance2011} when being initialized properly and therefore are our first choice.

The second order Taylor expansion of $f(X)$ around ${X}_{k}$ gives rise to the quadratic model function $m_k^{NR}(d)$ and the Newton-Raphson step $d_{k+1}$, where the Newton step $d_{k+1}$ is the minimizer of $m_k^{NR}(d)$ when $H\left(X_k\right)$ is positive definite:

\begin{equation}\label{Newton step Taylor series expansion}
\left\{\hphantom{A}
\begin{array}{rcccl}
f\left({X}_{k}+d\right) & \approx &m_k^{NR}(d) & =& f\left(X_k\right) + d^T g\left(X_k\right) + \dfrac{1}{2!} d^T H\left(X_k\right) d \\[12pt]
d_{k+1} &= &\arg\min\limits_{d\in \mathbb{R}^3} m_k^{NR}(d) & =& - H\left(X_k\right)^{-1} g\left(X_k\right)
\end{array}\right.
\end{equation}

Similarly, the first order Taylor expansion of $r(X)$~\eqref{residue vector function} around ${X}_k$ gives rise to the Gauss-Newton step $s_{k+1}$, the minimizer to another quadratic model function $m_k^{GN}(s)$ of $f(X)$~\eqref{reprojection error cost function} around $X_k$:
\begin{equation}\label{Gauss-Newton Taylor expansion}
\left\{\begin{array}{rccll}
r\left(X_k+s\right) & \approx & \hat r(s) & = r\left(X_k\right) + J\left(X_k\right) s \\[12pt]
m_k^{GN}(s) & =  &\hat r(s)^T \hat r(s)& = f\left(X_k\right) + 2 s^T g\left(X_k\right) + s^T \left(J\left(X_k\right)^T J\left(X_k\right)\right) s \\[12pt]
s_{k+1} & = & \arg\min\limits_{s\in \mathbb{R}^3}{m_k^{GN}}(s)& = -\left(J\left(X_k\right)^TJ\left(X_k\right)\right)^{-1} g\left(X_k\right)= -J\left(X_k\right)^\dag r\left(X_k\right)\\[12pt]
\end{array}\right.
\end{equation}

Levenberg-Marquardt algorithm is considered as a modification on $J^TJ$ in the Gauss-Newton iteration, or Gauss-Newton algorithm with trust region strategy on each step~\cite{Nocedal2006, Dennis:1996}.
\begin{equation}\label{Levenberg-Marquardt step}
\begin{array}{rcll}
p_{k+1} & = & -\left(J\left(X_k\right)^TJ\left(X_k\right)+\mu_k {{I}}\right)^{-1} g\left(X_k\right)
\end{array}
\end{equation}
The Levenberg-Marquardt algorithms we use are those from~\cite{LM-Yamoshita:2001, Fan-LM:2005, Nocedal2006, NumericalRecipes2007a, Fan-LM:2012}, with $\mu_k = \left\|r(X_k)\right\|_2^\delta($ $\delta\in(1,2))$ for the $\mu$~\eqref{Levenberg-Marquardt step} updating in every iteration and is relatively expensive.

\begin{algorithm}[!htp]
    \caption{Soft line search with Armijo backtracking~\cite{LineSearch2001}}
    \label{alg:iLineSearch}
    \begin{algorithmic}[1]
        \Procedure{SoftLineSearch}{$k, X_k, d_k, @f\left(X\right)$}\Comment{modified Armijo backtracking}
        \State $\gamma\gets \text{0.01}, \delta \gets \text{0.25} $ \Comment{set the line search parameters: $\gamma\in\left(0,0.5\right),\delta\in\left(0,1\right)$}
        \State $i \gets 0$ \Comment{so as to compatible with the underlying iteration}
        \Repeat \Comment{$a\bmod b$}
        \State $\alpha \gets \delta^i$
        \State $\alpha_k \gets \alpha$
        \If {$f(X_k+\alpha*d_k)\leq f(X_k) - \gamma * \alpha^3 * \left\|d_k\right\|_2^3$} \Comment{Armijo backtracking criterion}
        \State \textbf{return} $\alpha_k$  \Comment{return step length if meeting criterion}
        \EndIf
        \State {increment $i$ by 1}
        \Until{$i \ge 20$}\label{alg:iLSendRepeat}
       \State \textbf{return} $\alpha_k$\Comment{return after a max loop number}
       \EndProcedure
   \end{algorithmic}
\end{algorithm}

The two major iterative approaches we suggest to use are Gauss-Newton hybrid with globalizing strategies~\ref{alg:iLineSearch} and~\ref{alg:trustRegion}: global Gauss-Newton~\cite{Poliak1987, Dennis:1996, Nocedal2006, NumericalRecipes2007a}, denoted as {\em gGN} hereafter. The soft line search strategy with Armijo backtracking rule is as in algoirthm~\ref{alg:iLineSearch}, and simple trust region by Steihaug's method is as in algorithm~\ref{alg:trustRegion}, the theoretically {\em local} convergency (to critical points of $f(X)$) of which have been depicted and proven in literatures~\cite{Poliak1987, Dennis:1996, Nocedal2006, immoptibox2010}. Unless otherwise specified, {\em gGN} represents Gauss-Newton with~\ref{alg:iLineSearch} in the numerical experiments.

Since without any globalizing strategy, the underlying Newton-Raphson and Gauss-Newton iterative methods work both accurate and efficient for more than 99\% of the real cases, the globalizing strategies~\ref{alg:iLineSearch} and~\ref{alg:trustRegion} better be hybrid with underlying Gauss-Newton iteration in a {\em smooth} manner. For example, a major difference between the trust region~\ref{alg:trustRegion} version {\em gGN} and Levenberg-Marquardt is that trust region step-updating~\ref{alg:trustRegion} is only implemented when the new $f_{k+1}=f(X_{k+1})$ is greater than $f_0=f(X_0)$ in {\em gGN}, which makes the {\em gGN} more efficient without losing robustness. Too frequent trust region step-updating in Levenberg-Marquardt also ruins the accuracy according to our numerical experiments. Such hybridisation is also recommended to be used in the line search~\ref{alg:iLineSearch} version {\em gGN}.

\begin{algorithm}[!htp]
    \caption{Trust region algorithm: update $s_k$ by conjugate gradient method}
    \label{alg:trustRegion}
    \begin{algorithmic}[1]
        \Procedure{TrustRegion}{$g_k,B_k, X_k, @f\left(X\right)$}\Comment{simple trust region algorithm}
        \State $i\gets 0, x_i \gets X_k, \epsilon \gets 1.0\text{e}^{-8},\Delta_i \gets 1.0, \eta_s \gets 0.1, \eta_v \gets 0.9, \gamma_{\text{inc}}\gets 4, \gamma_{\text{red}} \gets 0.25$
        \Repeat
        \State {model function: $ m_i\left(s \right)\gets \left(-s^Tg_k- \dfrac{1}{2}s^TB_ks\right)$} \Comment{2nd order Taylor exp: $f\left(x_i\right)-f\left(x_i+s\right)$}
        \State {$s_i \gets \arg \min\limits_{\left\|s\right\|^2\le \Delta_i^2} m_i(s)$ }\Comment{solve subproblem by Steihaug method}\label{alg:trustRegionSubproblem}
        \State {$\rho_i \gets \dfrac{f\left(x_i\right)-f\left(x_i+s_i\right)}{m_i\left(s_i\right)}$}\Comment{The ratio of actual to predicted reduction}
        \If {$\rho_i \ge \eta_v$} \Comment{$m_k\left(s\right)$ approximates $f$ reduction very successful}
        \State {$x_{i+1} \gets x_{i}+s_i$}
        \State {$\Delta_{i+1}\gets \Delta_i * \gamma_{\text{inc}}$}\Comment{increase trust region radius $\Delta_i$}
        \ElsIf {$\rho_i \ge \eta_s$}  \Comment{$m_k\left(s\right)$ approximates $f$ reduction successful}
        \State {$x_{i+1} \gets x_{i}+s_i$}
        \State {$\Delta_{i+1}\gets \Delta_i $}
        \Else{}\Comment{\Comment{$m_k\left(s\right)$ does not approximate $f$ reduction} when $\rho_i < \eta_s $}
        \State {$x_{i+1}\gets x_i$}
        \State {$\Delta_{i+1}\gets \Delta_i * \gamma_{\text{red}}$}\Comment{reduce trust region radius $\Delta_i$}
        \EndIf
        \State {increment $i$ by 1}
       \Until {$\left\|g_i\right\|\le \epsilon$ \textbf{or} $i\ge 100$}
       \State \textbf{return} $s_k$
       \EndProcedure\label{alg:iTRendRepeast}
   \end{algorithmic}
\end{algorithm}
Numerical experiments indicate that if accurate derivative computation in section~\ref{section:symbolic-numeric derivatives} is used, all the Newton-like iterative methods initialized by {\em symmedian points}~\ref{suboptimal methods section} are L2 optimal~\cite{Hartley2013IJCV, Chesi2011tfml} for most real cases. 
Here we use four synthetic data examples from Chesi et al~\cite{Chesi2011tfml} to illustrate the L2 optimality of the iterative methods, and the conservative case of {\em tfml} does not occur for any of the iterative methods at all.

\vspace{0.2cm}
\textbf{Synthetic examples} The synthetic data examples are based on the four cameras defined as in~\eqref{ChesiExample01}. The ``SA2", ``SA3" and ``SA4" examples are the cases with the first 2, 3 and 4 cameras as in~\eqref{ChesiExample01} respectively and all their images are $(0,0,1)^T$. The conservative case ``Con", which {\em tfml} method fails in finding the optimal solution to,  has the first three cameras and its 2D images are: $x_1=\left(0.9,-0.9,1\right)^T,x_2=\left(0.6,2,1\right)^T,x_3=\left(2,1.3,1\right)^T$ respectively.

{\scriptsize\begin{equation}\label{ChesiExample01}
\hspace*{-0.15in}
\begin{array}{l}
P_1=\left[
\begin{array}{cccc}
 1 & 0 & 0 & 0 \\
 0 & 1 & 0 & 0 \\
 0 & 0 & 1 & 1 \\
\end{array}
\right], P_2=\left[
\begin{array}{cccc}
 -1 & -1 & -1 & 0 \\
 1 & 0 & -1 & 1 \\
 0 & 0 & 1 & 1 \\
\end{array}
\right],
P_3=\left[
\begin{array}{cccc}
 0 & -1 & 0 & 0 \\
 0 & 0 & -1 & 1 \\
 -1 & -1 & 0 & 1 \\
\end{array}
\right],
P_4=\left[
\begin{array}{cccc}
 0 & -1 & -1 & 0 \\
 0 & 1 & -1 & 1 \\
 1 & 0 & 1 & 1 \\
\end{array}
\right]
\end{array}
\end{equation}}
Comparison results are listed as in table~\ref{comparison table 1}.
\def\arraystretch{1.05}
{\small\begin{table}[!hpbt]
\begin{center}
\caption{{\normalsize Triangulation results comparison between {\em tfml} and iterative methods}}
\label{comparison table 1}
\advance\leftskip-.25cm
\begin{tabular}{ccrc}
\hlinewd{1.5pt}
\tc{Exmp.} & \tc{Method} & \tc{Triangulation result} & \tc{Reprojection error} \\[1.25ex]
\hline
\tc{\text{SA2}}&\tc{\text{\em tfml}}&\tr{\text{\small(-0.272727272727{\color{magenta}398},-0.18181818181{\color{magenta}7941},0.636363636363{\color{magenta}190})}}&\tc{\text{\small 0.055555555555556}}\\
\tc{\text{SA2}}&\tc{\text{\em gGN}} &\tr{\text{\small(-0.272727272727273,-0.181818181818182,0.636363636363636)}}&\tc{\text{\small 0.055555555555556}}\\
\tc{\text{SA3}}&\tc{\text{\em tfml}}&\tr{\text{\small(-0.3025060{\color{magenta}37933953},-0.160909{\color{magenta}286697078},0.7990907{\color{magenta}47348768})}}&\tc{\text{\small 0.10521103596214\color{magenta}3}}\\
\tc{\text{SA3}}&\tc{\text{\em gGN}} &\tr{\text{\small(-0.302506061882800,-0.160909312731383,0.799090767385097)}}&\tc{\text{\small 0.10521103596214\color{black}2}}\\
\tc{\text{SA4}}&\tc{\text{\em tfml}}&\tr{\text{\small(-0.232284{\color{magenta}343064664},-0.334519{\color{magenta}175175504},0.6968068{\color{magenta}78848929})}}&\tc{\text{\small 0.2099061662632\color{magenta}81}}\\
\tc{\text{SA4}}&\tc{\text{\em gGN}} &\tr{\text{\small(-0.232284268136407,-0.334519054968205,0.696806894375664)}}&\tc{\text{\small 0.2099061662632\color{black}48}}\\
\tc{\text{Con}}&\tc{\text{\em tfml}}&\tr{\text{\small(1.{\color{magenta}314094728910344},-1.{\color{magenta}106491029764633},0.{\color{magenta}043599248387159})}} &\tc{\text{\small 1.2\color{magenta}65349079248799}}\\
\tc{\text{Con}}&\tc{\text{\em gGN}} &\tr{\text{\small(1.424098078272550,-1.238341159147880,0.115482211291935)}} &\tc{\text{\small 1.2\color{black}23123745015136}}\\
\hline
\end{tabular}
\end{center}
\end{table}}
Since all the Newton-Raphson, Gauss-Newton and Levenberg-Marqquardt methods perform similar, we only list the global Gauss-Newton (`` gGN" with~\ref{alg:iLineSearch}) results for comparison. Table~\ref{comparison table 1} indicates that iterative methods are more accurate which also globally solves the conservative case for {\em tfml}. The L2 optimality can be easily verified by solving their~\eqref{nonlinear normal equation} via global optimization methods or per the criterion~\cite{Hartley2013IJCV} mentioned in section~\ref{section:criteria used}.

\subsection{Numeric criteria in evaluating triangulation solutions}\label{section:criteria used}

Because of their theoretically significance, there are global L2 optimality criteria developed by constructing the upper bound for $f(X)$ cost function or lower bound of its Hessian on a convex domain~\cite{Demidenko2000, Demidenko2006, Hartley2013IJCV} based on {\em sufficient} conditions of the convexity. A {\em necessary and sufficient} criterion naturally generated from {\em tfml} and {\em tpml} algorithms using the equality of $\mu_1$ and $\mu_2$ is therefore of special interests~\cite{Chesi2011tfml} though it is only limited to the proposed algorithms' verification. For those cases when camera number is small and when efficiency is not critical, it is also possible to compute all the real solutions to~\eqref{nonlinear normal equation} and compare the corresponding reprojection errors.

\begin{defi}[Numerical L2 optimality]\label{L2 optimal criterion}
A point is numerically L2 optimal if and only if it is a {\em good enough} approximate solution to the nonlinear normal equation \eqref{nonlinear normal equation} and its reprojection error is less than or equal to that of a  {\em nice} suboptimal estimation easy to obtain.
\end{defi}

The so-called {\em nice} suboptimal reprojection error can be the {\em upper local convexity level} as defined in~\cite{Demidenko2000, Demidenko2006} or simply use that of {\em symmedian point}, which works acceptable for most cases. The upper local convexity level is a sufficient criterion of L2 optimality but is rather difficult to compute accurately:
\begin{equation}\label{eqn:upper local convexity level}
f_{LC} \buildrel \Delta \over = \min\limits_{X\in \mathbb{R}^3}\min\limits_{y\in\mathbb{R}^3}\dfrac{y^TJ(X)^TJ(X)y}{\sum\limits_{i=1}^{2n}\left(y^T\nabla^2\phi_i(X) y\right)^2}
\end{equation}
A more favorable efficient L2 optimality verification approach with high success ratio is the sufficient criteria via investigating the lower bounds of Hessian of $f(X)$ on the convex intersection set of $n$ cone domains~\cite{Hartley2013IJCV}.

For the iterative methods proposed, we also use the following criteria to pre-determine whether the triangulation problem is a hard case or not, and the accuracy of a final solution.

Numerical experiments indicate that the square of an intrinsic curvature $\rho$ of $r(X)$ around a specific point $X$ works very well in picking out those hard cases, which is the reciprocal of the maximum eigen value $\lambda_{\text{max}}$ of symmetric matrix $K$~\eqref{eqn:curvatureK} determined by using $J^\dag$, the Moore-Penrose pseudo-inverse of $J$~\eqref{Jacobian of residue function}, and second order derivatives of $r(X)$~\cite{Dennis:1996, NewtonAffineInvariance2011}:
\begin{equation}\label{eqn:curvatureK}
K_{2n\times 2n}(X) \buildrel \Delta \over = -\left(J^\dag(X)\right)^T_{2n\times 3}\left(\sum\limits_{i=1}^{2n}\phi_i(X)\nabla^2\phi_i(X)\right)_{3\times 3}J^\dag(X)_{3\times 2n}
\end{equation}
The {\em intrinsic curvature} rule to determine the solvability via Gauss-Newton iteration of $\hat X^*$ is as~\cite{wedin:1977, Dennis:1996, NewtonAffineInvariance2011}:
\begin{equation}\label{eqn:averageResidueNorm}
\begin{array}{rcl}
\rho^2\left(\hat X^*\right) \buildrel \Delta \over = \dfrac{1}{\lambda^2_{\text{max}}\left( K\left(\hat X^*\right)\right)}&{\geqslant}& \gamma^2 \left(\hat X^*\right) \buildrel \Delta \over =\left\|r\left(\hat X^*\right)\right\|^2_2 =f\left(\hat X^*\right),\\[16pt]
K(X)\text{ is as in~\eqref{eqn:curvatureK},} &&f(X)\text{ is as in~\eqref{reprojection error cost function}, and }r(X)\text{ is as in~\eqref{residue vector function}}
\end{array}
\end{equation}

The the maximum absolute value of eigenvalues of $K$ indicates the local convergence rate of Gauss-Newton iteration; and the maximum eigenvalue $\lambda_{\text{max}}$ of $K$~\eqref{eqn:curvatureK},  is useful to determine whether a least square problem is easily solvable via Newton-like iteration from a specific initialization. 
A rule of thumb useful in determine the solvability of an L2 triangulation problem by Newton-like iterative method proposed is whether $\rho^2$ is significantly larger than $\gamma^2$ at the symmedian point $X_0$. The thumb rule also works for most of the global minimum determination when $K_{\text{LC}}$ is smaller than $\epsilon$ and the reprojection error at a point $\hat X$ is smaller than that of $X_0$. Equation~\eqref{eqn:averageResidueNorm} indicates that, the larger residue it has at the initializer the more difficult a triangulation problem is for iterative methods to solve because of its nonlinearity and multiple local minima, which is verified in our numerical experiments on extensive data sets~\cite{BundlerHugeDataECCV2010}.

A quantitative criterion for accuracy estimation inspired by Kantorovich theorem~\cite{NewtonIteration2008, NewtonAffineInvariance2011} is the 2-norm of the iterative step at the current $\hat X$:
\begin{defi}[LC distances]\label{LC Radius definition}
An estimation to the {\em local convergence} accuracy of $\hat X$,  Kantorovich distance $K_{\text{LC}}$, is defined as:
\begin{equation}\label{Kantorovich distance}
K_{\text{LC}} = \left\|H\left(\hat X\right)^{-1}g\left(\hat X\right)\right\|_2
\end{equation}
\end{defi}

$K_{\text{LC}}\leqslant$ $\epsilon\approx \sqrt{2.22\times 10^{-16}}\approx 1.49\times 10^{-8}$ for double precision computation usually means the current solution $\hat X$ is a numerically good enough critical point of $f(X)$.  The negative logarithm of $K_{\text{LC}}$ also approximately indicate the accuracy of convergence in significant decimal digits.

\section{Numerical Results for real data sets}\label{section:numerical examples}
Further numerical experiments are mainly conducted on the real data sets made available online by Oxford visual geometry group (\url{http://www.robots.ox.ac.uk/~vgg/data/data-mview.html}). Numerical results indicate that the L2 triangulation of all those data sets, {\em dinosaur, model house, corridor, Merton colleges I, II} and {\em III, University library} and {\em Wadham College}, can be globally solved by iteration methods~\eqref{Newton step Taylor series expansion} and~\eqref{Gauss-Newton Taylor expansion} in high efficiency with or without globalizing strategies~\ref{alg:iLineSearch} and \ref{alg:trustRegion}. Levenberg-Marquardt~\eqref{Levenberg-Marquardt step} only loses accuracy in very rare cases. The IEEE754 double precision C++ implementations of these iterative methods are conducted on a Windows\textsuperscript{\textregistered} computer with a 3.4GHz Intel\textsuperscript{\textregistered} i7 CPU.

Though global Gauss-Newton method({\em gGN}), i.e., iteration~\eqref{Gauss-Newton Taylor expansion} with Armijo backtracking line search strategy~\ref{alg:iLineSearch}, is relatively slower than Newton-Raphson~\eqref{Newton step Taylor series expansion}, it is more robust and therefore more favourable for general cases.

We present here comparison results between {\em gGN} and {\em tfml}~\cite{Chesi2011tfml} on their ACT(average computing time) and R.E.(reprojection error) for the following data sets only: 1) {\em dinosaur} in table~\ref{table:dinosaur}, which has the 21 camera case; 2) {\em corridor} in table~\ref{table:corridor}, which has the {\em tfml} conservative case (point No. 514); 3) {\em model house} in table~\ref{table:house}, which has the maximum percentage of more than 4 camera cases. All indicate the iterative methods  significantly outperform {\em tfml} in both efficiency and accuracy.

\def\arraystretch{1.1}
{\begin{table}[!hpbt]
\begin{center}
\caption{{\normalsize {\em dinosaur} data set results comparison between {\em tfml} and global Gauss-Newton method}}
\label{table:dinosaur}
\advance\leftskip-.25cm
\begin{tabular}{rrccrr}
\hlinewd{1.2pt}
\tc{$n$} & \tc{\# points} & \tc{ACT(s,{\em tfml}~\cite{Chesi2011tfml})} & \tc{ACT(s,{\em gGN})} & \tc{R.E.({\em tfml}~\cite{Chesi2011tfml})} & \tc{R.E.({\em gGN})} \\[1.2ex]
\hline
\tr{\text{2}} &\tr{\text{2300}}&\tc{\text{0.010}} &\tc{\text{0.0000362740}}&\tr{\text{233.8453557}}  &\tr{\text{233.8453557}}  \\
\tr{\text{3}} &\tr{\text{1167}}&\tc{\text{0.048}} &\tc{\text{0.0000408420}}&\tr{\text{8073.5262739}} &\tr{\text{8073.5262739}} \\
\tr{\text{4}} &\tr{\text{584}} &\tc{\text{0.060}} &\tc{\text{0.0000443918}}&\tr{\text{14972.8533254}}&\tr{\text{14972.8533254}}\\
\tr{\text{5}} &\tr{\text{375}} &\tc{\text{0.071}} &\tc{\text{0.0000477932}}&\tr{\text{4450.9466754}} &\tr{\text{4450.9466754}} \\
\tr{\text{6}} &\tr{\text{221}} &\tc{\text{0.080}} &\tc{\text{0.0000505274
}}&\tr{\text{10995.10677{\color{magenta}40}}}&\tr{\text{10995.10677{\color{black}39}}}\\
\tr{\text{7}} &\tr{\text{141}} &\tc{\text{0.097}} &\tc{\text{0.0000521147}}&\tr{\text{2955.318639{\color{magenta}2}}} &\tr{\text{2955.318639{\color{black}1}}} \\
\tr{\text{8}} &\tr{\text{88}}  &\tc{\text{0.115}} &\tc{\text{0.0000548824}}&\tr{\text{5396.742364{\color{magenta}7}}} &\tr{\text{5396.742364{\color{black}6}}} \\
\tr{\text{9}} &\tr{\text{44}}  &\tc{\text{0.148}} &\tc{\text{0.0000574861}}&\tr{\text{391.319527{\color{magenta}8}}}  &\tr{\text{391.319527{\color{black}7}}}  \\
\tr{\text{10}}&\tr{\text{26}}  &\tc{\text{0.175}} &\tc{\text{0.0000615819}}&\tr{\text{222.4767185}}  &\tr{\text{222.4767185}}  \\
\tr{\text{11}}&\tr{\text{15}}  &\tc{\text{0.215}} &\tc{\text{0.0000737776}}&\tr{\text{2930.4360{\color{magenta}400}}} &\tr{\text{2930.4360{\color{black}398}}} \\
\tr{\text{12}}&\tr{\text{14}}  &\tc{\text{0.270}} &\tc{\text{0.0000680938}}&\tr{\text{62.38270{\color{magenta}50}}}   &\tr{\text{62.38270{\color{black}49}}}   \\
\tr{\text{13}}&\tr{\text{5}}&\tc{\text{0.303}}&\tc{\text{0.0000799962}}&\tr{\text{250.05697{\color{magenta}20}}}&\tr{\text{250.05697{\color{black}19}}}\\
\tr{\text{14}}&\tr{\text{2}}&\tc{\text{0.390}}&\tc{\text{0.0000736568}}&\tr{\text{9.5944806}}&\tr{\text{9.5944806}}\\
\tr{\text{21}}&\tr{\text{1}}&\tc{\text{1.094}}&\tc{\text{0.0001041459}}&\tr{\text{28.4078252}}&\tr{\text{28.4078252}}\\
\hline
\tr{\text{Total}}&\tr{\text{4983}}&\tc{\text{203.614}}&\tc{\text{0.2051264558}}&\tr{\text{50973.01367{\color{magenta}74}}}&\tr{\text{50973.01367{\color{black}65}}}   \\
\hline
\end{tabular}
\end{center}
\end{table}}

{\begin{table}[!hpbt]
\begin{center}
\caption{{\normalsize {\em corridor} data set results comparison between {\em tfml} and global Gauss-Newton method}}
\label{table:corridor}
\advance\leftskip-.25cm
\begin{tabular}{rrccrr}
\hlinewd{1.2pt}
\tc{$n$} & \tc{\# points} & \tc{ACT(s,{\em tfml}~\cite{Chesi2011tfml})} & \tc{ACT(s,{\em gGN})} & \tc{R.E.({\em tfml}~\cite{Chesi2011tfml})} & \tc{R.E.({\em gGN})} \\[1.2ex]
\hline
\tr{\text{3}}&\tr{\text{341}}&\tc{\text{0.045}}&\tr{\text{0.0000616368}}&\tr{\text{94.4{\color{magenta}950818}}}&\tr{\text{94.4{\color{black}847897}}}\\
\tr{\text{5}}&\tr{\text{146}}&\tc{\text{0.078}}&\tr{\text{0.0000655332}}&\tr{\text{109.9579{\color{magenta}865}}}&\tr{\text{109.9579{\color{black}785}}}\\
\tr{\text{7}}&\tr{\text{88}}&\tc{\text{0.133}}&\tr{\text{0.0000808400}}&\tr{\text{135.1818{\color{magenta}406}}}&\tr{\text{135.1818{\color{black}014}}}\\
\tr{\text{9}}&\tr{\text{58}}&\tc{\text{0.220}}&\tr{\text{0.0000968958}}&\tr{\text{119.653{\color{magenta}4304}}}&\tr{\text{119.653{\color{black}3555}}}\\
\tr{\text{11}}&\tr{\text{104}}&\tc{\text{0.307}}&\tr{\text{0.0001014581}}&\tr{\text{204.112{\color{magenta}9585}}}&\tr{\text{204.112{\color{black}8521}}}\\
\hline
\tr{\text{Total}}&\tr{\text{737}}&\tc{\text{83.125}}&\tc{\text{0.0538715274}}&\tr{\text{663.{\color{magenta}4012979}}}&\tr{\text{663.{\color{black}3907772}}}\\
\hline
\end{tabular}
\end{center}
\end{table}}

{\begin{table}[!hpbt]
\begin{center}
\caption{{\normalsize {\em model house} data set results comparison between {\em tfml} and global Gauss-Newton method}}
\label{table:house}
\advance\leftskip-.25cm
\begin{tabular}{rrccrr}
\hlinewd{1.2pt}
\tc{$n$} & \tc{\# points} & \tc{ACT(s,{\em tfml}~\cite{Chesi2011tfml})} & \tc{ACT(s,{\em gGN})} & \tc{R.E.({\em tfml}~\cite{Chesi2011tfml})} & \tc{R.E.({\em gGN})} \\[1.2ex]
\hline
\tr{\text{3}}&\tr{\text{382}}&\tc{\text{0.056}}&\tr{\text{0.0000368079}}&\tr{\text{146.985{\color{magenta}7377}}}&\tr{\text{146.985{\color{black}6478}}}\\
\tr{\text{4}}&\tr{\text{19}}&\tc{\text{0.083}}&\tr{\text{0.0000921823}}&\tr{\text{23.2842{\color{magenta}602}}}&\tr{\text{23.2842{\color{black}599}}}\\
\tr{\text{5}}&\tr{\text{158}}&\tc{\text{0.073}}&\tr{\text{0.0000514310}}&\tr{\text{538.832{\color{magenta}9042}}}&\tr{\text{538.832{\color{black}8594}}}\\
\tr{\text{6}}&\tr{\text{3}}&\tc{\text{0.089}}&\tr{\text{0.0001322201}}&\tr{\text{15.38770{\color{magenta}83}}}&\tr{\text{15.38770{\color{black}45}}}\\
\tr{\text{7}}&\tr{\text{90}}&\tc{\text{0.109}}&\tr{\text{0.0000673846}}&\tr{\text{304.239{\color{magenta}3664}}}&\tr{\text{304.239{\color{black}2904}}}\\
\tr{\text{8}}&\tr{\text{1}}&\tc{\text{0.172}}&\tr{\text{0.0001234658}}&\tr{\text{7.534364{\color{magenta}2}}}&\tr{\text{7.534364{\color{black}1}}}\\
\tr{\text{9}}&\tr{\text{12}}&\tc{\text{0.185}}&\tr{\text{0.0001421567}}&\tr{\text{63.4833{\color{magenta}436}}}&\tr{\text{63.4833{\color{black}353}}}\\
\tr{\text{10}}&\tr{\text{7}}&\tc{\text{0.230}}&\tr{\text{0.0001360150}}&\tr{\text{11.209{\color{magenta}8018}}}&\tr{\text{11.209{\color{black}7303}}}\\
\hline
\tr{\text{Total}}&\tr{\text{672}}&\tc{\text{48.582}}&\tc{\text{0.03318089788}}&\tr{\text{1110.957{\color{magenta}4864}}}&\tr{\text{1110.957{\color{black}1917}}}\\
\hline
\end{tabular}
\end{center}
\end{table}}

First, from the tables~\ref{comparison table 1}~\texttildelow~\ref{table:house}, the conservative case for {\em tfml} never occurred for {\em gGN}. Note that the ``conservative case" for {\em tfml} occurs only in {\em corridor} data set (point No.514), all other results by {\em tfml} are {\em L2 optimal} per the criterion by Chesi et al~\cite{Chesi2011tfml}. By comparing the reprojection errors, it is easy to conclude that {\em gGN} results which are generally more accurate with smaller reprojection errors are also L2 optimal. The optimality of the {\em gGN} for the 3-view conservative case in {\em corridor} can be easily verified since which has been globally solven.

About efficiency, the three-view C++ implementation of {\em gGN} iteration are significantly faster than the C++ implementation of the three-view only L2 optimal methods~\cite{HowHard3view2005, Byroed07, Lu2007}; both efficiency and reprojection error of {\em gGN} are better than the the C++ implementation of the suboptimal methods by Recker et al~\cite{Recker2013WACV, Recker2013swarm}.

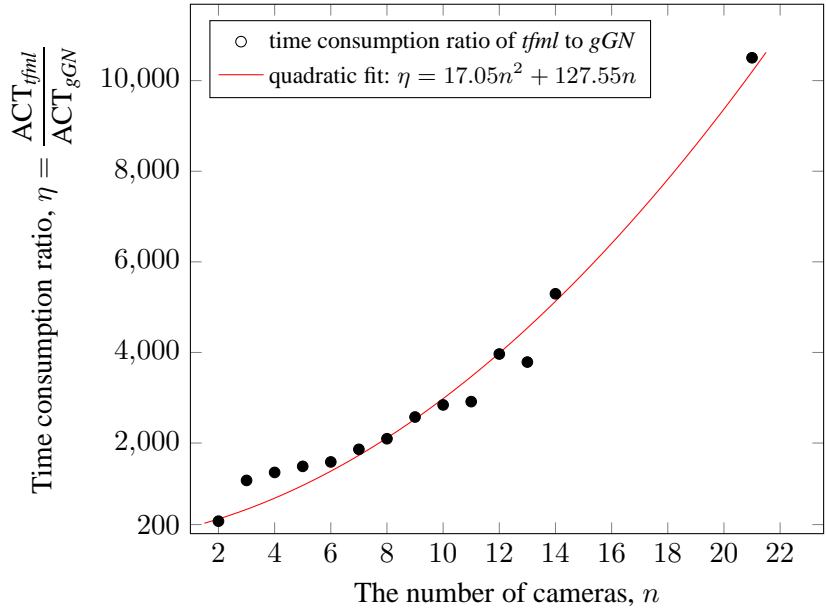
\begin{figure*}[!hpt]
\begin{center}
\begin{tikzpicture}
    \pgfplotsset{width=10cm,
    legend style={font=\footnotesize}}
    \begin{axis}[
    domain=1:22,
    xmin=1,
    xtick = {2,4,6,8,10,12,14,16,18,20,22},
    xminorgrids = true,
    ymin =0,
    xlabel={The number of cameras, $n$},
    ylabel={Time consumption ratio, $\eta =\dfrac{\text{ACT}_{\text{{\em tfml}}}}{\text{ACT}_{\text{\em gGN}}}$},
    legend cell align = left,
    legend pos = north west,
    ymin =0,
    ytick = {200,2000,4000,6000,8000,10000},
        y tick label style={
        /pgf/number format/.cd,
        fixed,
        fixed zerofill,
        precision=0,
        /tikz/.cd
    },
     yticklabel=\pgfmathprintnumber{\tick},
    scaled y ticks=base 10:0,
]
    \addplot[only marks] table[x =X,y =Y]{myData.dat};
    \addlegendentry{time consumption ratio of {\em tfml} to {\em gGN}};
    \addplot [domain=1.5:21.5,samples=50,red]({x},{17.05*x^2+127.55*x}); 
    \addlegendentry{quadratic fit: $\eta=17.05 n^2 +127.55 n $};
    \end{axis}
    \end{tikzpicture}
\caption{The trend of {\em tfml} to {\em gGN} time consumption ratio versus camera number (Oxford dinosaur data set)}
\label{fig:efficiency ratio trend}
\end{center}
\end{figure*}

There is a trend of the ACT ratio $\eta(n) = \dfrac{\text{ACT}_\text{{\em tfml}}}{\text{ACT}_\text{\em gGN}}$ between {\em tfml} and {\em gGN}: {\em tfml} becomes slow faster than {\em gGN} because its EVP scale is getting larger with the increase of camera number~\cite{Chesi2011tfml}, as is illustrated in figure~\ref{fig:efficiency ratio trend}. For the 21-view case, {\em gGN}(C++) is more than 10000 times faster than {\em tfml}(per ACT in~\cite{Chesi2011tfml}).

Extensive numerical experiments are carried out based on the data sets by Agarwal et al~\cite{BundlerHugeDataECCV2010}, where radial distortions of the calibrated cameras are neglected for the purpose of algorithm verification. Iterative method {\em gGN} has only achieved L2 optimality for 99.7\% of the points since there exist large residue cases or outliers. However, globalizing strategies~\ref{alg:iLineSearch} and~\ref{alg:trustRegion} assure local convergence to critical points and significant reporjection error improvement of the {\em symmedian point} initializers for all those hard cases. And in such hard cases, neither iterative methods, nor {\em tfml} has absolute advantage over their peers; while {\em gGN} is the most favourable method which has the overall robustness, high efficiency, higher success ratio of convergence to critical points and highest ratio of achieving the lowest reprojection error in such extensive numerical experiments. %

\section{Discussion and Conclusion}\label{discussion}

By symmedian point initialization and accurate computation of derivatives, Newton type iterative methods can solve most of the multiple view L2 triangulation problems both efficiently and accurately, which means the difficulty of the multiple local minima of the nonconvex reprojection error cost function $f(X)$ can be easily overcome in such real cases.

This indicate that {\em symmedian points} can efficiently locate the attraction basin of the optimal solution to $f(X)$ in most real cases which simplifies the multiple view L2 triangulation problem into convex ones,  and accurate computation of derivatives are critical for Newton-like methods to be successful in solving multiple view L2 triangulation problem.

In order to handle those hard cases where the nonlinearity of $f(X)$ is so high and reprojection error is large at the initializers, globalizing strategies~\ref{alg:iLineSearch} and~\ref{alg:trustRegion} are proposed to use smoothly-hybrid with the underlying Gauss-Newton iteration, which outperform Levenberg-Marquardt and other methods in robustness and efficiency, achieving high success ratio of convergence to critical points and significant reprojection error improvement over the symmedian point initializers.

This means bundle adjustment with appropriate implementations can significantly outperform its peers in solving optimal triangulation problems.

Similar to what has been proposed in~\cite{Hartley2013IJCV}, in the rare cases where {\em symmedian point} triangulation fails to locate the optimal solution attraction basin it is usually because the point has large noise, in which case in a large-scale reconstruction problem, the best option is probably to remove the point from consideration.

Future work on optimal triangulation may focus on improving initialization technique which assures to locate the attraction basin of the global minimum, while the problems of L2 optimality guaranteed triangulation for multiple view cases continue to be NP-hard with no simple solution in general~\cite{Hartley2013IJCV}. And it is useful to develop efficient and reliable strategies, similar to the intrinsic normal curvature~\eqref{eqn:averageResidueNorm} for Gauss-Newton iterations, so as to previously determine whether a problem is solvable or not iteratively. 

\section*{Acknowledgement}

{\small
\bibliographystyle{ieee}
\bibliography{triangulation}
}

\section*{Supplementary Materials}
{\footnotesize
\begin{enumerate}[A. ]
\item  Newton-like triangulator source codes in visual C++ for Windows platforms.
\item  Oxford visual geometry group data sets selected.
\end{enumerate}}
\end{document}